\documentclass{article}

\usepackage{microtype}
\usepackage{amsmath}
\usepackage{amsfonts}
\usepackage{graphicx}
\usepackage{caption}
\usepackage{subcaption}
\usepackage{booktabs} 
\usepackage{csquotes}
\usepackage{siunitx}
\usepackage{dcolumn}
\usepackage{float}

\usepackage{hyperref}

\usepackage{changepage}
\usepackage{placeins}

\newlength{\offsetpage}
{\end{adjustwidth}}

\usepackage[accepted]{icml2018}

\include{macros}

\icmltitlerunning{Provably Minimally-Distorted Adversarial Examples}

\begin{document}

\twocolumn[
  \icmltitle{Provably Minimally-Distorted Adversarial Examples}



\icmlsetsymbol{equal}{*}

\begin{icmlauthorlist}
\icmlauthor{Nicholas Carlini}{equal,ucb}
\icmlauthor{Guy Katz}{equal,st}
\icmlauthor{Clark Barrett}{st}
\icmlauthor{David L. Dill}{st}
\end{icmlauthorlist}

\icmlaffiliation{st}{Stanford University}
\icmlaffiliation{ucb}{University of California, Berkeley}

\icmlcorrespondingauthor{Guy Katz}{guyk@cs.stanford.edu}
\icmlcorrespondingauthor{Nicholas Carlini}{npc@berkeley.edu}

\icmlkeywords{Machine Learning, ICML}

\vskip 0.3in
]



\printAffiliationsAndNotice{\icmlEqualContribution} 

\begin{abstract}
The ability to deploy neural networks in real-world, safety-critical
systems is severely limited by the presence of \emph{adversarial
  examples}: slightly perturbed inputs that are misclassified by the
network.
In recent years, several techniques have been proposed for increasing
robustness to adversarial examples --- and yet most of these have been
quickly shown to be vulnerable to future attacks. For example, over half of the
defenses proposed by papers accepted at ICLR 2018 have already been broken.
%
We propose to address this difficulty through formal verification
techniques. We show how to construct \emph{provably minimally distorted adversarial
examples}: 
given an arbitrary neural network and input sample,
we can construct adversarial examples which we prove are of minimal
distortion.
Using this approach, we demonstrate
that one of the recent ICLR defense proposals,
adversarial retraining, provably succeeds at increasing the distortion
required to construct adversarial examples by a factor of $4.2$.

\end{abstract}

\section{Introduction}
While machine learning, and neural networks in particular, have seen significant
success, recent work~\cite{szegedy2013intriguing} has shown that an adversary can cause unintended behavior
by performing slight modifications to the input at test-time. In neural
networks used as classifiers,
these \emph{adversarial examples} are produced by taking some normal instance
that is classified correctly, and applying a slight perturbation to cause it to be
misclassified as any target desired by the adversary. This phenomenon,
which has been shown to affect most state-of-the-art networks, poses a
significant hindrance to deploying neural networks in safety-critical settings.

Many effective techniques have been proposed for generating adversarial examples
\cite{szegedy2013intriguing,goodfellow2014explaining,moosavi2015deepfool,
  carlini2016towards}; and, conversely, several techniques have been
proposed for training networks that are more robust to these examples
\cite{huang2015learning,zheng2016improving,hendrik2017detecting,
  hendrycks2017early,madry2017towards}. Unfortunately,
it has proven difficult to accurately assess the
robustness of any given defense by evaluating it
against existing attack techniques.
In several cases, 
a defensive technique that was at first
thought to produce robust networks was later shown to be susceptible
to new kinds of attacks.
Most recently, at ICLR 2018, seven accepted defenses were shown
to be vulnerable to attack \cite{athalye2018obfuscated}.
This ongoing cycle thus cast a doubt in any
newly-proposed defensive technique.

In recent years, new techniques have been proposed for the \emph{formal
  verification} of neural
networks~\cite{katz2017reluplex,PuTa10,PuTa12,HuKwWaWu16,Ehlers2017}. These
techniques take a network 
and a desired property, and formally \textbf{prove} that the network satisfies
the property --- or provide an input for which the property is
violated, if such an input exists. Verification techniques can be used
to find adversarial examples for a given input
point and some allowed amount of distortion, but they tend to be
significantly slower than gradient-based
techniques~\cite{katz2017reluplex,PuTa12,katz2017fvav}.

\textbf{Contributions.}
In this paper we propose a method for using formal verification
to assess the effectiveness of adversarial example attacks and defenses.
The key idea  apply verification to construct
\emph{provably minimally distorted adversarial examples}: inputs that are misclassified
by a classifier but
are provably minimally distorted under a chosen distance metric.
We perform two forms of analysis with this approach.

\begin{itemize}
\item \emph{Attack Evaluation.} We use provably
  minimally distorted adversarial examples to evaluate the efficacy
of a recent attack \cite{carlini2016towards} at generating adversarial examples, and
find it produces produces adversarial examples within $12\%$ of optimal
on our small model on the MNIST dataset.
This suggests that iterative optimization-based attacks are indeed
effective at generating adversarial examples, and strengthens the hypothesis of
\citet{madry2017towards} that first-order attacks are ``universal''.

\item \emph{Defense Evaluation.} More interestingly, we can also apply this technique to prove properties
about defenses. Given an arbitrary defense, we can apply it to a small
enough problem amenable to verficiation, and prove properties about
that defense in the restricted setting.
As a case study, we evaluate the robustness of adversarial training
as performed by \citet{madry2017towards} at
defending against adversarial examples on the MNIST dataset.
This defense was \emph{emperically}
found to be among the strongest submitted to
ICLR 2018 \cite{athalye2018obfuscated}, and in this paper
we formally \emph{prove} this defense is effective: it succeeds at
increasing robustness to adversarial examples by $4.2\times$ on the
samples we examine.
While this does not gaurantee efficacy at larger scale, it does
gaurantee that, at least on small networks, this defense has
not just caused current attacks to fail; it has 
successfully managed to increase the robustness of neural networks
against \emph{all} future attacks.
To the best of our knowledge, we are the first to apply formal
verification to formally prove properties about defenses initially
designed with only emperical results.
\end{itemize}

\section{Background and Notation}
\label{sec:background}

\bigskip \noindent
\textbf{Neural network notation.} We regard a neural network as a function $F(\cdot)$ consisting
of multiple layers $F = F_n \circ F_{n-1} \circ \dots \circ F_1 \circ F_0$.
In this paper we exclusively study feed-forward neural networks used
for classification, and so the final layer $F_n$ is the softmax activation
function.
We refer to the output of the second-to-last-layer of the network (the 
input to $F_n$) as the logits and denote
this as $Z = F_{n-1} \circ \dots \circ F_1 \circ F_0$.
We define $\ell_F(x, y)$ to be the cross-entropy loss of the network $F$ on instance $x$ with
true label $y$.

We focus here on networks for classifying greyscale MNIST images.
 Input images
with width $W$ and height $H$ are represented
as points in the space $[0,1]^{W \cdot H}$.

\bigskip \noindent
\textbf{Adversarial examples.} \cite{szegedy2013intriguing}
Given an input $x$, classified originally as target $t = \text{arg max} F(x)$,
and a new desired target $t' \ne t$, we
call $x'$ a \emph{targeted adversarial example} if $\text{arg max} F(x') = t'$ and $x'$ is close
to $x$ under some given distance metric. 

Exactly which distance metric to use to properly evaluate the ``closeness''
between $x$ and $x'$ is a difficult question
\cite{rozsa2016adversarial,xiao2018spatially,zhao2018generating}. 
However, almost all work in this space has decided on using $L_p$ distances
to measure distortion \cite{szegedy2013intriguing,goodfellow2014explaining,moosavi2015deepfool,hendrik2017detecting,carlini2016towards,madry2017towards}, and
Every defense at ICLR 2018 argues $L_p$ robustness.
We believe that considering more sophisticated distance metrics is
an important direction of research, but for consistency with prior work,
in this paper we evaluate using the $L_\infty$ and $L_1$ distance metrics.

\bigskip \noindent
\textbf{Generating adversarial examples.}
We make use of three popular methods for constructing adversarial examples:
\begin{enumerate}
\item The \emph{Fast Gradient Method} (\emph{FGM}) \cite{goodfellow2014explaining}
  is a one-step algorithm that takes a single
  step in the direction of the gradient.
  \[ x' = \text{FGM}(x) = \text{clip}_{[0,1]}(x + \epsilon \text{sign}(\nabla \ell_F(x, y))) \]  
  where $\epsilon$ controls the step size taken, and $\text{clip}$ ensures that
  the adversarial example resides in the valid image space from $0$ to $1$.
\item The \emph{Basic Iterative Method} (\emph{BIM}) \cite{kurakin2016adversarial}
  (sometimes also called \emph{Projected Gradient Descent} \cite{madry2017towards})
  can be regarded as an iterative application of the fast gradient method.
  Initially it lets $x'_0=x$ and then uses the update rule
  \[ x'_{i+1} = \text{clip}_{[x-\alpha,x+\alpha]}(FGM(x'_i)) \]
  Intuitively, in each iteration this attack takes a step of size
  $\epsilon$ as per
  the FGM method, but it iterates this process while keeping each $x'_i$ within the
  $\alpha$-sized ball of $x$.
\item The \emph{Carlini and Wagner} (\emph{CW})
\sloppy
  \cite{carlini2016towards} method is an iterative attack that
  constructs adversarial examples by approximately solving the minimization
  problem $\min d(x,x')$ such that $F(x') = t'$ for the 
  attacker-chosen target
  $t'$, where $d(\cdot)$ is an appropriate distance metric.
  Since the constrained optimization is difficult, instead they choose to solve
  $\min d(x,x') + c \cdot g(x')$ where $g(x')$ is a loss function that
  encodes how close $x'$ is to being adversarial. Specifically, they set
  \[ g(x') = \max(\max \{ Z(x')_i : i \ne t\} -
  Z(x')_t, 0). \]
  $Z(\cdot)$, the logits of the network, are used instead of the softmax output
  because it was found to provide superior results.
  Although it was originally constructed to optimize for $L_2$ distortion, we
  use it with $L_1$
and $L_\infty$
 distortions in this paper.
   
\end{enumerate}
 
\bigskip \noindent
\textbf{Neural network verification.}
 The intended use of deep neural networks as controllers in
safety-critical
systems~\cite{JuLoBrOwKo16,BoDeDwFiFlGoJaMoMuZhZhZhZi16}
has sparked an interest in developing techniques for verifying
that they satisfy various properties~\cite{PuTa10, PuTa12, HuKwWaWu16, Ehlers2017, katz2017reluplex}.
Here we focus on the recently-proposed Reluplex
algorithm~\cite{katz2017reluplex}: a simplex-based
approach that can effectively tackle networks with piecewise-linear
activation functions, such as rectified linear units (ReLUs) or
max-pooling layers. Reluplex is known to be sound and complete, and
so it is suitable for establishing adversarial examples of provably minimally distortion.

In~\cite{katz2017reluplex} it is shown that Reluplex can be used 
to determine whether there exists an adversarial example within
distance $\delta$ of some input point $x$. This is performed by
encoding the neural network itself and the constraints regarding
$\delta$ as a set of linear equations and ReLU constraints, and then 
having Reluplex attempt to prove the property that ``there does not
exist an input point within distance $\delta$ of $x$ that is assigned
a different label than $x$''.
 Reluplex either responds that the property holds, in which case there is
 no adversarial example within distance $\delta$ of $x$, or it returns
 a counter-example which constitutes the sought-after adversarial input.
By invoking Reluplex iteratively and
applying binary search, one can approximate the optimal $\delta$
(i.e., the largest $\delta$ for which no adversarial example exists)
up to a desired precision~\cite{katz2017reluplex}.

The proof-of-concept implementation of Reluplex described
in~\cite{katz2017reluplex} supported only networks with the ReLU
activation function, and could only handle the $L_\infty$ norm as a
distance metric. Here we use a simple encoding that allows us to use
it for the $L_1$ norm as well. 

\bigskip \noindent
\textbf{Adversarial training.} Adversarial training is perhaps the first proposed
defense against adversarial examples \cite{szegedy2013intriguing},
and is a conceptually straightforward
approach. The defender
trains a classifier, generates adversarial examples for that
classifier, retrains
the classifier using the
adversarial examples, and repeats.

Formally, the defender attempts to solve the following formulation
$$\theta^* = \mathop{\text{arg min}}_\theta\; \mathbb{E}_{x \in \mathcal{X}} \left[ \max_{\delta \in [-\epsilon,\epsilon]^N} \ell(x+\delta; F_\theta) \right]$$
by approximating the inner minimization step with an existing
attack technique.

Recent work has shown \cite{madry2017towards}
that for networks with sufficient capacity, adversarial training can be
an effective defense even against the most powerful attacks today
by training against equally powerful attacks.

It is known that if adversarial training is performed using
weaker attacks, such as the fast gradient sign method, then it is still
possible to construct adversarial examples
by using stronger attacks \cite{tramer2018ensemble}.

It is an open question whether
adversarial training using stronger attacks (such as PGD)
will actually increase robustness to all attacks,
or whether such training 
will be effective at preventing only current attacks.

\bigskip \noindent
\textbf{Provable (certified) defenses.}
Very recent work at ICLR 2018 has begun to construct certified defenses to
adversarial examples. These defenses can give a proof of robustness that
adversarial examples of distortion at most $\epsilon$ cause a test loss of
at most $\delta$.
This work is an extremely important direction of research that applies
formal verification to the process of constructing provably sound defenses
to adversarial examples.
However, the major drawback of these approaches is that certified
defenses so far can \emph{only} be applied to small networks on small
datasets.

In contrast, this work can take an \emph{arbitrary} defense (that can
be applied to networks of any size), and formally
prove properties about it on a small dataset.
If a defense is not effective on a large dataset, it is also likely
to be ineffective on the small dataset we study, and we will therefore
be able to show it is not effective.

Put differently,
our work shares the same key limitation of certified defenses:
when scaling to larger datasets, we are no longer able to offer
provably gaurantees.
However, because the defenses we study scale to larger datasets, even
though our proofs of robustness do not, it is still possible to apply
these defenses with increased confidence in their security.

\section{Model Setup}
\label{sec:modelSetup}

The problem of neural network verification that we consider here is an NP-complete
problem~\cite{katz2017reluplex}, and despite recent progress only
networks with a few hundred nodes can be soundly verified. Thus,
in order to evaluate our approach we trained a small network 
over the MNIST data set. This network is a fully-connected, 3-layer
network that achieves a $97\%$ accuracy despite having only 20k weights and
consisting of fewer than 100 hidden neurons (24 in each layer).
As verification of neural networks becomes more scalable in the
future, our approach could become applicable to larger networks and
additional data sets.

\begin{table*}[]
\centering
\caption{Evaluating our technique on the MNIST dataset}
\begin{tabular}[htp]{lc|crlrlrlr}
  \toprule
  &&& \multicolumn{1}{c}{Number}
  && \multicolumn{1}{c}{Carlini-}
  && \multicolumn{1}{c}{Minimally Distorted}
  && \multicolumn{1}{c}{Percent} \\
  &&& \multicolumn{1}{c}{of Points}
  && \multicolumn{1}{c}{Wagner}
  && \multicolumn{1}{c}{Adversarial Example}
  && \multicolumn{1}{c}{ Improvement}
    \\
  \midrule
  $N$, $L_\infty$       &&& 38/90  && 0.042  && 0.038 && 11.632 \\
                       &&& 90/90  && 0.063 && 0.061 && 6.027 \\
  \midrule
  $N$, $L_1$           &&& 6/90   && 1.94  && 1.731 && 34.909 \\
                       &&& 90/90  && 7.551 && 7.492 && 3.297 \\
  \midrule
  $\bar{N}$, $L_\infty$ &&& 81/90  && 0.211  && 0.193 && 11.637 \\
                       &&& 90/90  && 0.219  && 0.203 && 10.568 \\
  \midrule
  $\bar{N}$, $L_1$     &&& 64/90  && 6.44 && 6.36 && 6.285 \\
                       &&& 90/90  && 8.187 && 8.128 && 4.486 \\
  \bottomrule
\end{tabular}%
\label{table:comparison}
\end{table*}%

For verification, we use the proof-of-concept implementation of
Reluplex available online~\cite{reluplexCode}. 
The only non-linear operator that this implementation was originally 
designed to support is the ReLU function, but we observe that it can
support also $\max$ operators 
using the following encoding:
\[
\max(x,y) = \text{ReLU}(x-y) + y
\]
This fact allows the encoding of max operators using ReLUs, and
consequently to encode max-pooling layers 
into Reluplex (although we did not experiment with such 
layers in this paper). 
Thus, it allows us to extend the results from~\cite{katz2017reluplex} and
measure distances with the $L_1$ norm as well as the $L_\infty$
norm, by encoding absolute values using ReLUs:
\[
|x| = \max(x,-x) = \text{ReLU}(2x) - x
\]
Because the $L_1$ distance between two points is defined as a sum of
absolute values, this encoding allowed us to encode $L_1$ distances into Reluplex
without modifying its code.
We point out, however, that an increase in the number of ReLU constraints in the
input adversely affects Reluplex's performance. For example, in the
case of the MNIST dataset, encoding $L_1$
distance entails adding a ReLU constraint for each input of the 784
input coordinates.
It is thus not surprising that experiments using $L_1$ typically took longer to
finish than those using $L_\infty$.
 
Each individual experiment that we conducted included a network $F$,
a distance metric $d\in \{L_1,L_\infty\}$, an input point $x$, a
target label $\ell' \neq F(x)$, and an initial adversarial input $x'_{init}$ for which
$F(x'_{init})=\ell'$. The goal of the experiment was then to find minimally distorted example
$x_{\ell'}$, such that $F(x_{\ell'})=\ell'$ and $d(x,x_{\ell'})$ is minimal. As
explained in Section~\ref{sec:background}, this is performed by
iteratively invoking Reluplex and performing a binary search. 


Intuitively, $\delta_{max}$ indicates the distance to the closest adversarial
input currently known, and the provably minimally-distroted input is known to be
in the range between $\delta_{min}$ and $\delta_{max}$. Thus,
$\delta_{max}$ is initialized using the distance of the initial
adversarial input provided, and $\delta_{min}$ is initialized to 0.
 The search procedure iteratively
shrinks the range $\delta_{max}-\delta_{min}$ until it is below a certain
threshold (we used $10^{-3}$ for our experiments). It then returns
$\delta_{max}$ as the distance to the provably minimally distorted
adversarial example, and this is guaranteed to
be accurate up to the specified precision. The provably minimally distorted input
itself is also returned.

For the initial $x'_{init}$ in our experiments we used
an adversarial input found using the CW
attack. We note that Reluplex invocations are computationally expensive, and so it is
better to start with $x'_{init}$ as close as possible to $x$, 
in order to reduce the number of required iterations until
$\delta_{max}-\delta_{min}$ is sufficiently small. For the same reason,
experiments using the $L_1$ distance metric were slower than those
using $L_\infty$: the initial distances 
were typically much larger, which required additional iterations.

\section{Evaluation}
\label{sec:evaluation}

For evaluation purposes we arbitrarily selected 10 source images with
known labels from the MNIST test set.
 We considered two neural networks --- the one described in
Section~\ref{sec:modelSetup}, denoted $N$, and also a version of $N$
that has been trained  with adversarial training as described
in~\cite{madry2017towards}, denoted $\bar{N}$. We also considered two
distance metrics, $L_1$ and $L_\infty$. For every combination of neural
network, distance metric and labeled source image $x$, we considered each
of the 9 other possible labels for $x$. For each of these we used the CW
attack to produce an initial targeted adversarial example, and then used 
Reluplex to search for a
provably minimally distorted example. The results are given in
Table~\ref{table:comparison}.


Each major row of the table corresponds to specific neural network and
distance metric (as indicated in the first column), and
describes 90 individual experiments (10 inputs, times 9 target labels for each
input). The first sub-row within each row considers just those
experiments for which
Reluplex terminated successfully, whereas the
second sub-row considers all 90 examples, including those where
Reluplex timed out. Whenever
a timeout occurred, we considered the last (smallest)
$\delta_{max}$ that was discovered by the search before it timed out
as the algorithm's output. The other
columns of the table indicate the average distance to the adversarial
examples found by the CW attack, the average distance to the
minimally-distorted adversarial examples found by our technique, and the
average improvement rate of our technique over the CW attack.

Below we analyze the results in order to draw conclusions regarding
the CW attack and the defense of~\cite{madry2017towards}. While these
results naturally hold only for the networks we used and the inputs we
tested, we believe they provide some intuition as to how well the
tested attack and defense techniques perform. We intend to make our
data publicly available, and we encourage others to 
(i) evaluate new attack techniques using the minimally-distorted examples
  we have already discovered, and on additional ones; and
(ii) to use this approach for evaluating new defensive techniques.

\subsection{Evaluating Attacks}

\bigskip \noindent
\textbf{Iterative attacks produce near-optimal adversarial examples.}
As is shown by Table~\ref{table:comparison},
the adversarial examples produced by the CW attack are on average
within $11.6\%$ of the minimally-distorted example when using the $L_\infty$ norm,
and within $6.2\%$ of the minimally-distorted example when using $L_1$ (we consider
here just the terminated experiments, and ignore the $N,L_1$ category
where too few experiments terminated to draw a meaningful conclusion).
In particular, iterative attacks perform substantially better than
single-step methods, such as the fast gradient method. This is an 
expected result and is not surprising: the fast gradient method was
designed to show the linearity of neural networks, not to produce
high-quality adversarial examples.

This result supports the hypothesis of \citet{madry2017towards} who
argue first-order attacks (i.e., gradient-based methods) are ``universal''.
Further, this therefore justifies using first-order methods as the basis
of adversarial training; at least on the datasets we consider.

\bigskip \noindent
\textbf{There is still room for improving iterative attacks.}
Even on this very small and simple neural network, we observed that in many
instances the ground-truth adversarial example has a $30\%$
or $40\%$ lower distortion rate than the best iterative adversarial example.
The cause for
this is simple: gradient descent only finds a local minimum, not
a global minimum.

We have found that if we take a small step from
the original image \emph{in the direction of the minimally-distorted
  adversarial example}, then
gradient descent will converge to the minimally-distorted adversarial
example. Taking random steps and then performing gradient descent does
not help significantly.

\bigskip \noindent
\textbf{Suboptimal results are correlated.}
We have found that when the iterative attack performs suboptimally
compared to the minimally-distorteds example for one target label, it will often
perform poorly for many other target labels as well. These instances
are not always of larger absolute distortion, but a larger relative
gap on one instance often indicates that the relative gap will be
larger for other targets. For instance, on the adversarially trained
network attacked under $L_\infty$ distance, the ground-truth
adversarial examples for the digit $9$ were from $21\%$ to $47\%$
better than the iterative attack results.

\begin{figure}[H]
\begin{subfigure}{.5\textwidth}
  \hspace*{1cm}
  \begin{tabular}{llllllllll}
        \multicolumn{10}{c}{Reluplex, $N$, $L_\infty$, Target Label} \\
        0\, & 1\, & 2\, & 3\, & 4\,\, & 5\, & 6\, & 7\, & 8\, & 9\, \\
  \end{tabular}  \\
  {\rotatebox[origin=l]{90}{
      \begin{tabular}{llllllllll}
        \multicolumn{10}{c}{Source Label} \\
        9\, & 8\, & 7\, & 6\, & 5\,\, & 4\, & 3\, & 2\, & 1\, & 0\, \\
  \end{tabular}}}
  \centering
  \includegraphics[scale=0.125]{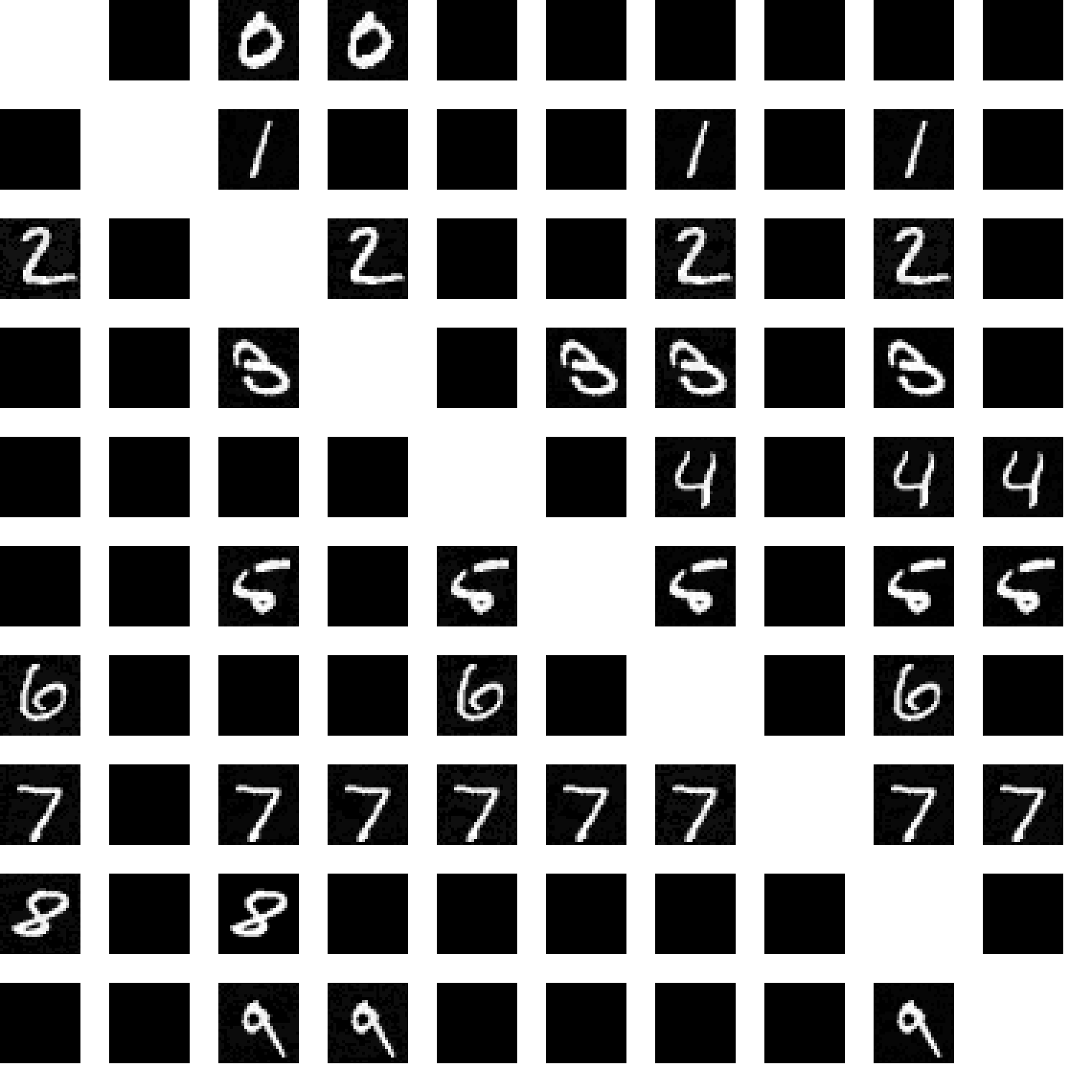}
  \caption{Adversarial Examples generated on a neural \\ network using
  Reluplex.}
\end{subfigure}
\begin{subfigure}{.5\textwidth}
  \hspace*{1cm}
  \begin{tabular}{llllllllll}
        \multicolumn{10}{c}{CW, $N$, $L_\infty$, Target Label} \\
        0\, & 1\, & 2\, & 3\, & 4\,\, & 5\, & 6\, & 7\, & 8\, & 9\, \\
  \end{tabular}  \\
  {\rotatebox[origin=l]{90}{
      \begin{tabular}{llllllllll}
        \multicolumn{10}{c}{Source Label} \\
        9\, & 8\, & 7\, & 6\, & 5\,\, & 4\, & 3\, & 2\, & 1\, & 0\, \\
  \end{tabular}}}
  \centering
  \includegraphics[scale=0.125]{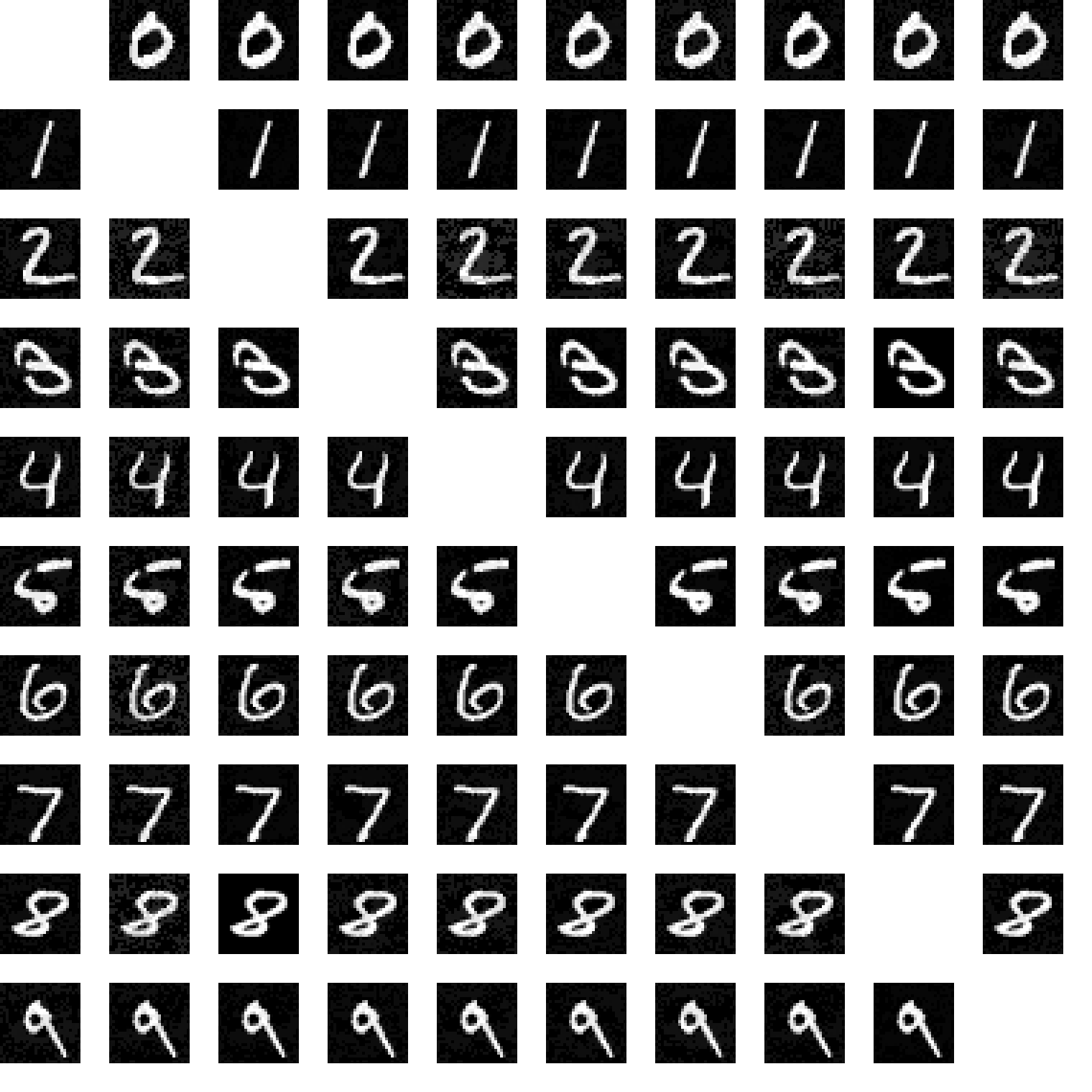}
  \caption{Adversarial Examples generated on a neural \\ network using
  \citet{carlini2016towards}.}
\end{subfigure}
\end{figure}

When we examined the most extreme cases in which this phenomenon was
observed, we found that, similarly to the case described above, the
large gap was caused by gradient descent initially leading
\emph{away} from the minimally-distorted example for most targets, resulting in
the discovery of an inferior, local minimum.

\subsection{Evaluating Defenses}

For the purpose of evaluating the defensive technique
of~\cite{madry2017towards}, we compared the $N,L_\infty$ and
$\bar{N},L_\infty$ experiments (the $L_1$ experiments were disregarded
because of the small number of experiments that terminated for the
$N,L_1$ case). Specifically, we compared the $N,L_\infty$ and $\bar{N},L_\infty$ experiments on the
subset of $35$ instances that terminated for both experiments. The
results appear in Table~\ref{table:comparison2}.

\begin{table}
\centering
\caption{
  Comparing the 35 instances on which Reluplex terminated
  for both $N,L_\infty$ and $\bar{N},L_\infty$.
}
\begin{tabular}[htp]{l|rrrr}
  \toprule
  & \multicolumn{1}{c}{Number}
  & \multicolumn{1}{c}{CW}
  & \multicolumn{1}{c}{Minimally}
  & \multicolumn{1}{c}{Percent}
    \\
  & \multicolumn{1}{c}{of Points}
  & \multicolumn{1}{c}{}
  & \multicolumn{1}{c}{Distorted}
  & \multicolumn{1}{c}{Improvement}
    \\
  \midrule
  $N$, $L_\infty$        & 35/35  & 0.042 & 0.039 & 12.319 \\
  $\bar{N}$, $L_\infty$  & 35/35  & 0.18  & 0.165 & 11.153 \\
\end{tabular}%
\label{table:comparison2}
\end{table}%

\bigskip \noindent
\textbf{The defense of \citet{madry2017towards} is effective.}
Our evaluation suggests that adversarial retraining is indeed effective:
it improves the distance to the minimally distorted adversarial examples by an average of 
423\% 
(from an average of 0.039 to an average of 0.165)
 on our small network.

Another interesting observation is that while adversarial retraining improves
the overall situation, we found several points in which it actually
made things worse --- i.e., the minimally distorted adversarial examples for the hardened network
were smaller than that of the original network. This behavior was
observed for 7 out of the 35 aforementioned experiments, with the average percent of 
degradation being 12.8\%. This seems to highlight the necessity of 
evaluating the effectiveness of a defensive technique, and the
robustness of a network in general, over a large dataset of
points. The question of how to pick a ``good'' set of points that
would adequately represent the behavior of the network remains open.

\bigskip \noindent
\textbf{Training on iterative attacks does not overfit.}
Overfitting is a an issue that is often encountered when performing
adversarial training. By this we mean that 
 a defense may overfit to the type of attack
used during training. When this occurs, the hardened network will have
high accuracy against the one attack used during training, but give low accuracy
on other attacks. We have found no evidence of overfitting when
performing the adversarial training of~\cite{madry2017towards}:
the minimally distorted adversarial examples improve on the CW attack by $12\%$ on both the
hardened and untrained networks.

\begin{figure}[H]
\begin{subfigure}{.5\textwidth}
  \hspace*{1cm}
  \begin{tabular}{llllllllll}
        \multicolumn{10}{c}{Reluplex, $\bar{N}$, $L_\infty$, Target Label} \\
        0\, & 1\, & 2\, & 3\, & 4\,\, & 5\, & 6\, & 7\, & 8\, & 9\, \\
  \end{tabular}  \\
  {\rotatebox[origin=l]{90}{
      \begin{tabular}{llllllllll}
        \multicolumn{10}{c}{Source Label} \\
        9\, & 8\, & 7\, & 6\, & 5\,\, & 4\, & 3\, & 2\, & 1\, & 0\, \\
  \end{tabular}}}
  \centering
  \includegraphics[scale=0.125]{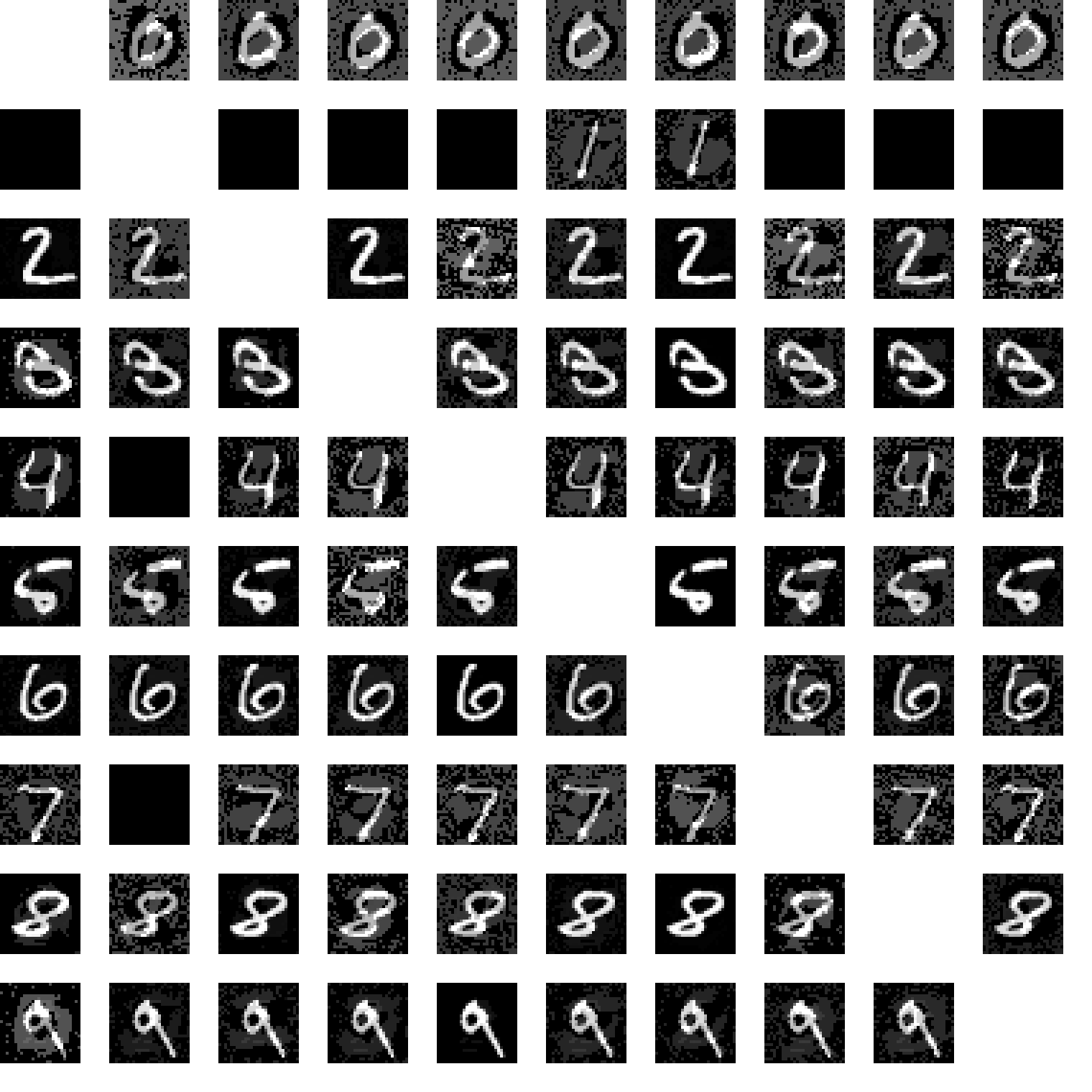}
  \caption{Adversarial Examples generated on \citet{madry2017towards} \\ using
  \citet{carlini2016towards}.}
\end{subfigure}
\begin{subfigure}{.5\textwidth}
  \hspace*{1cm}
  \begin{tabular}{llllllllll}
        \multicolumn{10}{c}{CW, $\bar{N}$, $L_\infty$, Target Label} \\
        0\, & 1\, & 2\, & 3\, & 4\,\, & 5\, & 6\, & 7\, & 8\, & 9\, \\
  \end{tabular}  \\
  {\rotatebox[origin=l]{90}{
      \begin{tabular}{llllllllll}
        \multicolumn{10}{c}{Source Label} \\
        9\, & 8\, & 7\, & 6\, & 5\,\, & 4\, & 3\, & 2\, & 1\, & 0\, \\
  \end{tabular}}}
  \centering
  \includegraphics[scale=0.125]{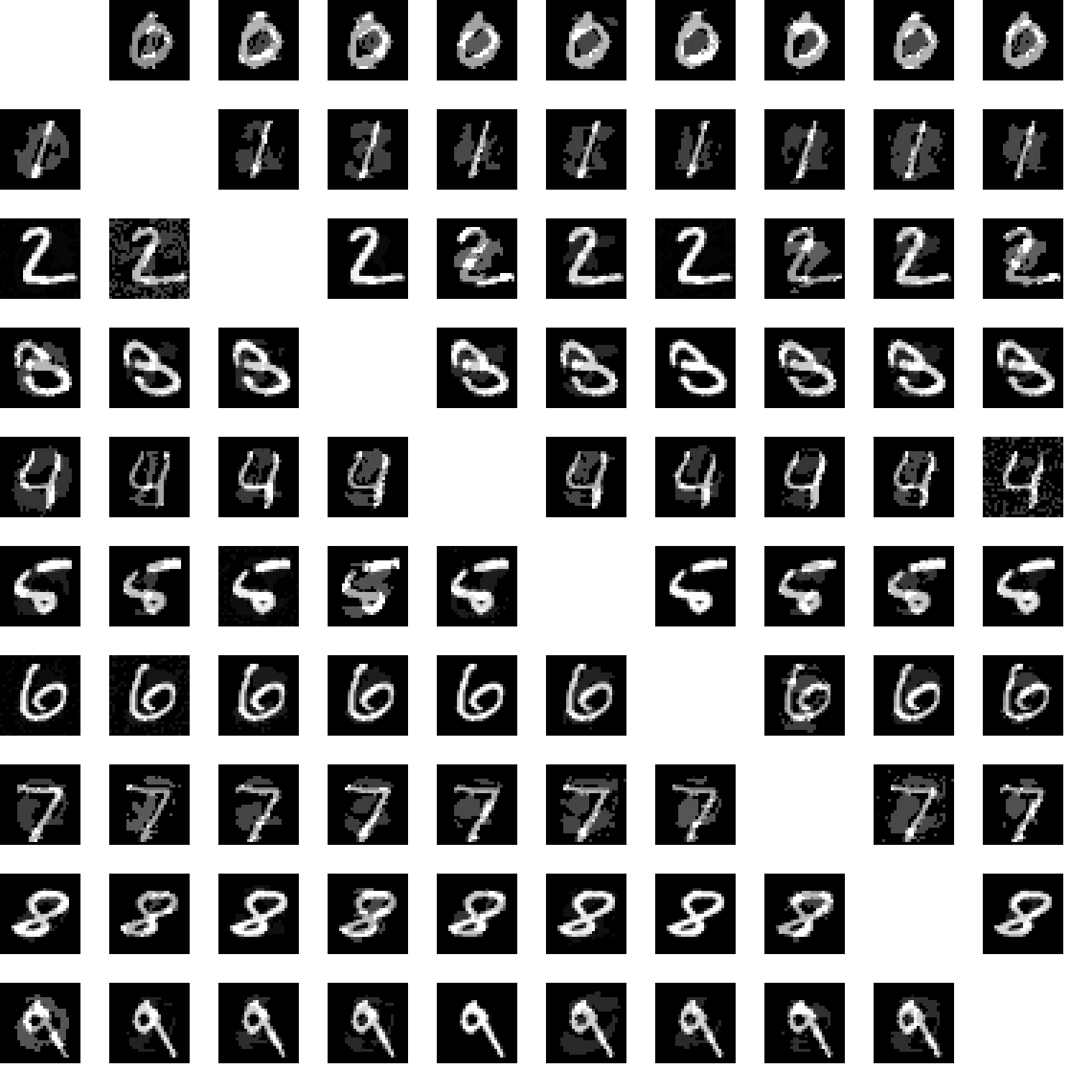}
  \caption{Adversarial Examples generated on \citet{madry2017towards} \\ using
  \citet{carlini2016towards}.}
\end{subfigure}
\end{figure}


 
\bigskip \noindent
\textbf{It is easier to formally analyze \citet{madry2017towards}.}
For both the $L_\infty$ and $L_1$ distance metrics, it
seems significantly easier to analyze the robustness of the adversarially
trained network: when using $L_\infty$, Reluplex terminated on 81 of
the 90 instances on the adversarially trained network, versus 38 on the
standard network; and for $L_1$, the termination rate was 64 for the
hardened network compared to just 6 on the standard network. 
We are still looking into the reason for this behavior. Naively, one
might assume that because the initial adversarial examples $x'_{init}$
provided to Reluplex have larger distance for
the hardened network, that these experiments will take longer to
converge --- but we were seeing an opposite behavior.

One possible explanation could be that the adversarially trained network
makes less use of the nonlinear ReLU units, and is therefore more
amenable to analysis with Reluplex. We empirically verify that this is
not the case.
For a given instance, we track, for each ReLU unit in the network, whether it is
in the saturated zero region, or the linear $x=y$ region. We then compute
the nonlinearity of the network as the number of units that change from the
saturated region to the linear region, or vice versa, when going from
the given input to the discovered adversarial example.
We find  that there is no statistically significant difference between the
nonlinearity of the two networks.

\section{Conclusion}
\label{sec:conclusion}

Neural networks hold great potential to be used in
safety-critical systems, but their susceptibility to adversarial
examples poses a significant hindrance. While defenses can be argued
secure against existing attacks, it is difficult to assess vulnerability
to future attacks.
The burgeoning field of neural network verification can
mitigate this problem, by allowing us to obtain an absolute measurement of
the usefulness of a defense, regardless of the attack to be used
against it.

In this paper, we introduce provably minimally distorted adversarial examples and show
how to construct them with formal verification approaches. We evaluate
one recent attack \cite{carlini2016towards} and find it often produces
adversarial examples whose distance is within
$6.6\%$ to $13\%$ of optimal, and one defense \cite{madry2017towards},
and find that it increases distortion
to the nearest adversarial example by an average of $423\%$ on the
MNIST dataset for our tested networks. To the best of our knowledge,
this is the first proof of robustness increase for a defense that was
not designed to be proven secure.

Currently available verification tools afford limited scalability,
which means experiments can only be conducted on small
networks. However, as better verification techniques are developed,
this limitation is expected to be lifted. Orthogonally, when
preparing to use a neural network in a safety-critical setting, users
may choose to design their networks as to make them particularly
amenable to verification techniques --- e.g., by using specific
activation functions or network topologies --- so that strong
guarantees about their correctness and robustness may be obtained.

\medskip
\noindent

{\footnotesize
\bibliographystyle{abbrvnat}
\bibliography{paper}
}

\newpage

\end{document}